\documentclass{article}

\usepackage[preprint]{tackling_climate_workshop_style}

\usepackage[utf8]{inputenc} 
\usepackage[T1]{fontenc}    
\usepackage{hyperref}       
\usepackage{url}            
\usepackage{booktabs}       
\usepackage{amsfonts}       
\usepackage{nicefrac}       
\usepackage{microtype}      
\usepackage{xcolor}         
\usepackage{graphicx} 
\usepackage{subfig}
\usepackage{subcaption}
\usepackage{xcolor}
\usepackage{makecell}

\title{Tree Species Classification using Machine Learning and 3D Tomographic SAR - a case study in Northern Europe}

\author{%
   Jumpei Takami\\
  United Nations Office \\ for Outer Space Affairs 
  \And
   Grace Colverd\\
  University of Cambridge \\
  \And
   Laura Schade\\
Department for Energy \\ Security and Net Zero - UK \\
  \AND
   Karol Bot \\
INSEC TEC\\
  \And
   Joseph A. Gallego-Mejia\\
Drexel University \\
\texttt{joseph.gallegomejia@drexel.edu}
}

\begin{document}

\maketitle

\begin{abstract}
Tree species classification plays an important role in nature conservation, forest inventories, forest management, and the protection of endangered species. Over the past four decades, remote sensing technologies have been extensively utilized for tree species classification, with Synthetic Aperture Radar (SAR) emerging as a key technique. In this study, we employed TomoSense, a 3D tomographic dataset, which utilizes a stack of single-look complex (SLC) images, a byproduct of SAR, captured at different incidence angles to generate a three-dimensional representation of the terrain. Our research focuses on evaluating multiple tabular machine-learning models using the height information derived from the tomographic image intensities to classify eight distinct tree species. The SLC data and tomographic imagery were analyzed across different polarimetric configurations and geosplit configurations. We investigated the impact of these variations on classification accuracy, comparing the performance of various tabular machine-learning models and optimizing them using Bayesian optimization. Additionally, we incorporated a proxy for actual tree height using point cloud data from Light Detection and Ranging (LiDAR) to provide height statistics associated with the model’s predictions. This comparison offers insights into the reliability of tomographic data in predicting tree species classification based on height.

\end{abstract}

\section{Introduction and Methodology}
Tree classification is critical for preserving forests, protecting endangered species, and assessing carbon sequestration, among other ecological benefits \cite{FASSNACHT201664}. This study leverages 3D tomographic images derived from Synthetic Aperture Radar (SAR) data, generated by TomoSense—an experiment funded by the European Space Agency (ESA)  \cite{tebaldini_tomosense_2023}. These images have undergone comprehensive preprocessing, including co-registration, geometric correction, image alignment, and noise reduction. In anticipation of the ESA's Biomass Satellite launch next year, which will operate in full tomographic mode for an entire year, this work explores the potential of tomographic images for accurate tree species classification. We utilized a manually classified dataset comprising eight distinct tree species within our area of interest, as outlined in Table \ref{tab:forest_types_with_percentages}, to serve as our ground truth.

To achieve optimal classification results, we applied a range of tabular machine learning models, utilizing tabular data and enhancing performance through Bayesian optimization combined with AutoML techniques. This approach enabled us to efficiently explore the model space and optimize performance metrics simultaneously. This work establishes a framework for future large-scale forest inventories, potentially enhancing forest management, biodiversity assessments, and carbon stock estimations.

\subsection{Scope and Related Work}
Tree species classification in forests typically involves a combination of morphological, physiological, and genetic methods to accurately identify and categorize different species \cite{zhang2023morphological}. Morphological methods rely on visible characteristics such as leaf shape, bark texture, tree height, and branching patterns. These features are often complemented by phenological observations, including flowering and fruiting times. Physiological methods can include analyzing the biochemical properties of tree tissues, such as leaf pigments or resin composition. Additionally, genetic methods, such as DNA barcoding and molecular markers, provide precise identification by analyzing the genetic material of the trees. In practice, these methods are often used in conjunction to improve accuracy, with field observations typically serving as the first step, followed by laboratory analysis when necessary. Advanced technologies, like remote sensing and machine learning, as employed in this work, are increasingly being integrated into species classification to enhance efficiency and precision, as well as enable global analysis \cite{pu2021mapping}.

The main stakeholders involved in tree species classification in forests, as in \cite{pelyukh2021stakeholder} include forestry professionals, conservationists, government agencies, researchers, and local communities. Forestry professionals and government agencies rely on accurate classification for sustainable forest management, policy-making, and regulatory compliance. Conservationists use species data to prioritize areas for protection and restoration, aiming to preserve biodiversity. Researchers are involved in advancing scientific understanding of forest ecosystems, contributing to climate change models and ecological studies. Local communities, especially indigenous groups, ar stakeholders as they often possess traditional knowledge of species and rely on forests for their livelihoods, making their involvement essential for culturally and ecologically sensitive management practices.

Synthetic Aperture Radar (SAR) data is increasingly being used to address the challenges of tree species classification in forests due to its ability to provide high-resolution, all-weather, and day-and-night imaging capabilities. Traditional methods of species classification, while effective, are often limited by accessibility, weather conditions, and the sheer scale of forested areas. SAR data overcomes these limitations by penetrating cloud cover and capturing detailed surface information, including tree structure and biomass, which are critical for distinguishing between species \cite{ren2024self}.  This remote sensing technology allows for more comprehensive and frequent monitoring of forests, making it a valuable tool for stakeholders who require accurate and timely data for conservation, forest management, and ecological research. By integrating SAR data with other classification methods, the accuracy and efficiency of species identification are significantly enhanced, providing a robust solution to the challenges posed by traditional techniques.

In the context of utilizing SAR data for tree species classification in forests, AutoGluon \cite{erickson_autogluon-tabular_2020} plays a role by automating the machine learning process, making it easier to develop and deploy highly accurate predictive models. AutoGluon is an open-source library that simplifies the process of applying machine learning by automatically selecting the best models and hyperparameters based on the input data, including complex SAR datasets. When applied to SAR data for tree species classification, AutoGluon can efficiently handle large and diverse datasets, selecting the best models that can extract relevant features such as texture, structure, and backscatter characteristics from the radar imagery. Among the models that AutoGluon might select and optimize are gradient boosting machines (e.g., XGBoost), convolutional neural networks (CNNs), and random forests, which are known for their strong performance in handling high-dimensional data and capturing complex patterns. These models, once optimized by AutoGluon, can significantly improve the accuracy of species classification by effectively leveraging the unique properties of SAR data, thereby enhancing decision-making in forestry management and conservation efforts.

\section{Experimental Design}
In this section, we explain the dataset's characteristics, the features used for creating the classification model, and the geographical split of the data.

\subsection{Dataset}
The TomoSense dataset provides several valuable resources, including single-look complex (SLC) images and point-cloud data captured from Light Detection and Ranging (LiDAR), among other useful information. For our experiments, we focused on the radar signal intensity derived from the constructed tomographic image. This tomographic data has a spatial resolution of 2 meters, with dimensions of 326 pixels in the range direction, 840 pixels in the azimuth direction, and 36 pixels in the height direction. Additionally, it includes three different polarimetric configurations (HH, HV, VV) and two opposite heading directions (North-West and South-East).

The species classification data collected by the field survey was provided by  Landesbetrieb Wald und Holz Nordrhein-Westfalen, Nationalparkforstamt Eifel, Germany. The biotope types group layer can be selected in the Web Map Service Landscape Information Collection (WMS LINFOS), which shows the spatial location of biotope types worthy of protection in North Rhine-Westphalia. Work was done to convert this data from an HTML format to a vector-based one that could be integrated with the TomoCube data. This dataset includes 8 species of tree (e.g Aspen, Pine). The dataset statistics are given in Appendix \ref{dataset_info}. The dataset is imbalanced, with the majority class (Aspen) accounting for 60\% of the study area.



\subsection{Machine Learning Features}

\begin{figure}
    \centering
        \caption{From left to right: the first image shows the single-look complex (SLC) image of the study area; the second depicts the intensity of the Tomographic SAR image; the third presents the cloud points from the Light Detection and Ranging (LiDAR) data; and the final image displays the tree species map for the study area.}
    \label{fig:intensity}
    \includegraphics[width=1\linewidth]{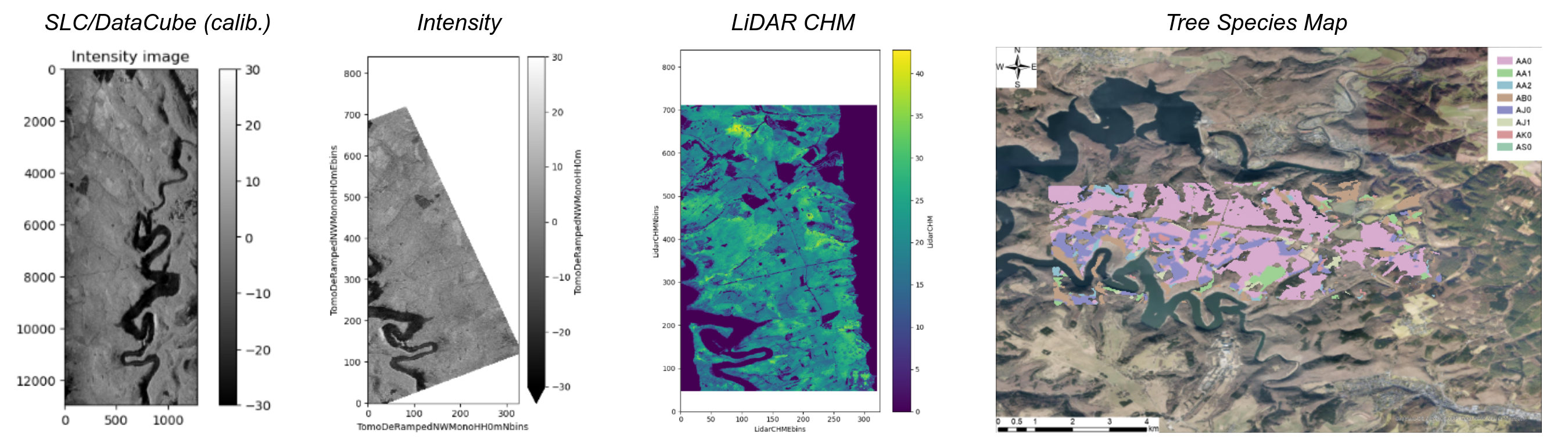}
\end{figure}

The tomographic SAR image used in our study has a 2-meter resolution along the elevation axis and a 1-meter resolution along both the azimuth and range axes. The data captures heights ranging from -10 meters to 50 meters. To structure our analysis, we employed a voxelized tomographic representation with a 1-meter by 1-meter resolution in the azimuth-range plane and a 2-meter resolution along the elevation axis.

As illustrated in Fig. \ref{fig:model}, the data processing pipeline involves several key steps. Initially, the original data is divided using one of two distinct spatial splits. Following this, the dataset is further separated by polarimetry, and we also evaluated combinations of the different polarimetric configurations. Additionally, we assessed whether incorporating spatial coordinates could enhance model performance. This process results in a feature array that includes the intensity values from the tomographic image, each associated with its respective polarimetric channel. Given that each heading captures only a portion of the total spatial area, as depicted in Fig. \ref{fig:intensity}, it is crucial to combine the data from both headings for a comprehensive analysis.

\subsection{Geographical Split}

\begin{figure}
    \centering
        \caption{Geographic data splits used for training and testing: The left figure illustrates the swath split, while the right figure depicts the square split.}
    \label{fig:geosplit}
    \includegraphics[width=0.45\linewidth]{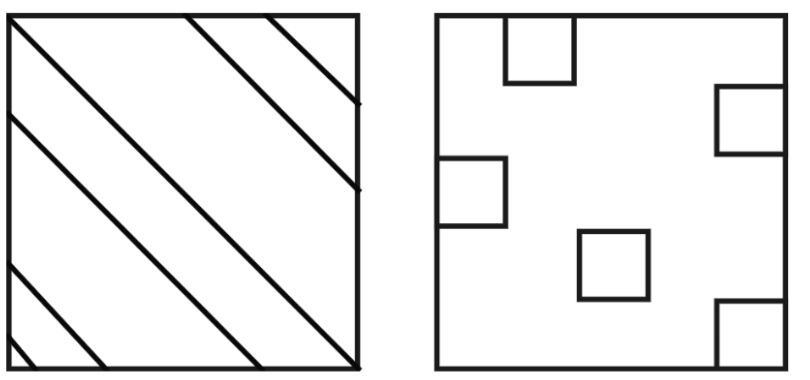}
\end{figure}

Spatial autocorrelation is a significant challenge in remote sensing, often leading to violations of the independence and identically distributed (i.i.d.) assumption during data splitting. Completely eliminating this issue remains a considerable challenge \cite{SALAZAR2022109885}. In this study, we addressed this problem by implementing two distinct data splitting strategies, as depicted in Fig. \ref{fig:geosplit}. These strategies include a swath split and randomly generated square regions. The swath split covers 20\% of the data's width, while the squares are set to 5\% of the width. For both swath and square splits, the data was divided into an 80:20 ratio for training and testing. The test data is held out until the final model is selected using AutoGluon, after which it is evaluated.

\subsection{Model Training}

In our study, we utilized AutoGluon (\citet{erickson_autogluon-tabular_2020}), an AutoML tool, to streamline the process of evaluating and selecting the best machine-learning models for tree species classification. AutoGluon simplifies the machine-learning pipeline by automatically tuning hyperparameters, selecting features, and optimizing models. Among the various models tested using AutoGluon, ensemble methods such as gradient boosting machines and deep neural networks stood out in terms of accuracy and performance. These models achieved high accuracy and robust performance, making them ideal for handling the complexities of TomoSAR data.

\section{Results and Discussion}
The comparisons of different geographic splitting for model species classification accuracy are given in Table \ref{tab:geo_split_res}.  
The increase in accuracy when the inputs include the range and azimuth per pixel (x,y) indicates that spatial information significantly enhances tree species identification. This improvement suggests that tree species distribution patterns have a strong spatial component, likely reflecting the clustering of similar species, perhaps due to topographic variations or competitive interactions. 
 
\begin{table}[]
\centering 
\caption{Geo-split and geo-coded comparisons for P band data.}
\label{tab:geo_split_res}
 
\begin{tabular}{@{}lllll@{}}
\toprule
Band & Geosplit & XY      & Accuracy & Balanced Accuracy \\ \midrule
P    & Square   & With XY & 0.77     & 0.26              \\
P    & Swathe   & With XY & 0.72     & 0.36              \\
P    & Square   & No XY   & 0.56     & 0.20              \\
P    & Swathe   & No XY   & 0.60     & 0.25              \\ \bottomrule
\end{tabular}
\end{table}


\begin{table}[h]
\caption{Classification Report: Class-wise Metrics (left) and Overall Metrics (right)
}
\label{tab:class_report}
\centering
\begin{tabular}{@{}p{0.5\linewidth}p{0.5\linewidth}@{}}
\begin{tabular}{@{}lllll@{}}
\toprule
Class & Precision & Recall & F1-Score & Support \\ \midrule
1     & 0.79      & 0.90   & 0.84     & 13529   \\
2     & 0.31      & 0.26   & 0.28     & 698     \\
3     & 0.28      & 0.17   & 0.21     & 679     \\
4     & 0.69      & 0.59   & 0.63     & 1919    \\
5     & 0.59      & 0.44   & 0.50     & 3318    \\
6     & 0.05      & 0.02   & 0.03     & 470     \\
7     & 0.18      & 0.10   & 0.13     & 125     \\
8     & 0.58      & 0.44   & 0.50     & 312     \\ \bottomrule
\end{tabular}
&
\begin{tabular}{@{}lll@{}}
\toprule
Metric       & Value & Support \\ \midrule
Accuracy     & 0.72  & 21050   \\
Macro Avg    & 0.39  & 21050   \\
Weighted Avg & 0.70  & 21050   \\ \bottomrule
\end{tabular}
\end{tabular}

\end{table}

\begin{figure}
    \centering
    \includegraphics[width=\linewidth]{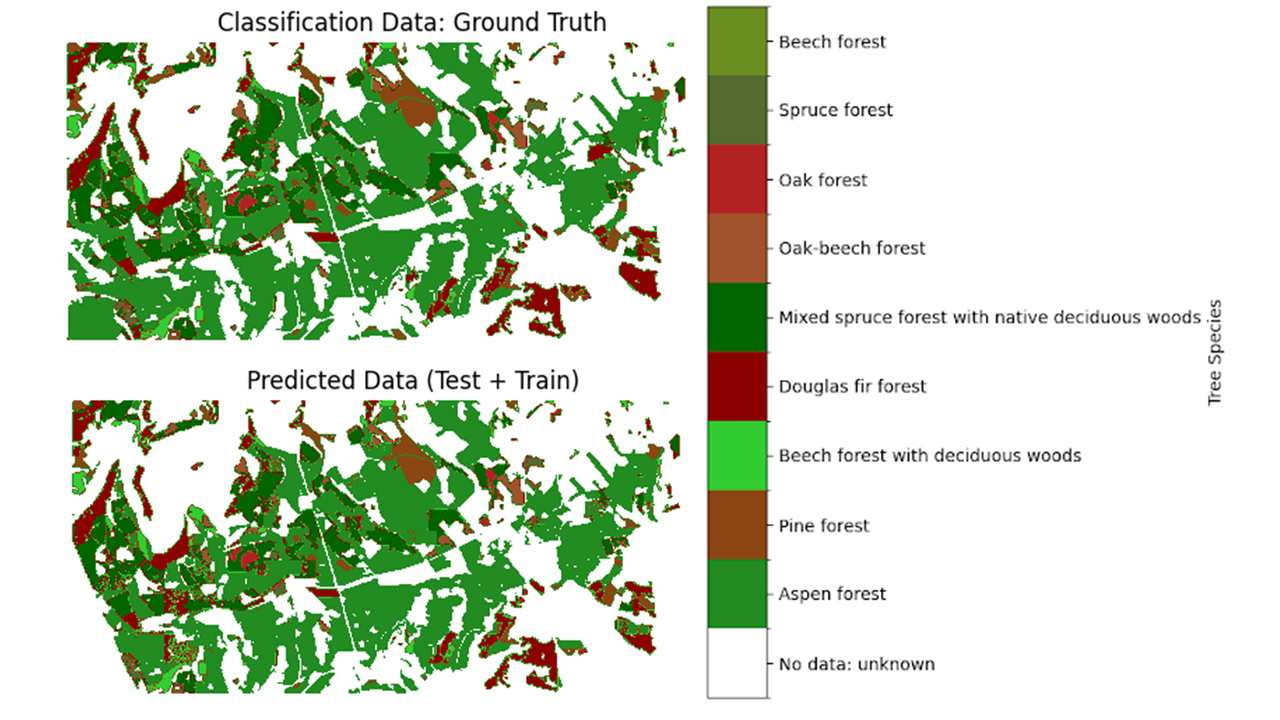}
    \caption{Result for Classification and Prediction}
    \label{fig:classification-map}
\end{figure}

In Table \ref{tab:class_report}, we presented a table with the classification report, including the precision, recall, F1-score, and support for each tree class, as well as the overall accuracy, macro average, and weighted average. In Figure \ref{fig:classification-map}, we display at the top the true tree classification and bottom the predicted classification.

The classification results show that some discrimination of species is possible with TomoSar data, and also the challenges in dealing with a highly imbalanced dataset. While the model achieves a reasonable overall accuracy of 72\%, this metric is heavily influenced by its strong performance on the dominant class (Class 1 Aspen, 64.3\% of samples, F1-score: 0.84). The substantial gap between the macro-average F1-score (0.39) and the weighted-average F1-score (0.70) underscores the model's struggle with minority classes, particularly Classes 2, 3, 6, and 7 (F1-scores < 0.30). This disparity suggests potential overfitting to the majority class and insufficient learning from underrepresented classes. The generally higher precision compared to recall for minority classes indicates a conservative prediction approach, possibly due to the class imbalance. 
Future research directions should focus on improving minority class prediction and investigating data augmentation or transfer learning to enhance the model's performance on underrepresented classes without compromising its accuracy on the majority class.

\section{Conclusions}
In this work, we investigate the potential of 3D  SAR images for tree species classification. Utilizing the TomoSense dataset, we transform the 3D image data into a tabular format, where each pixel corresponds to a specific height within the canopy. Our experiments focus on the P-band and examine how different polarimetries influence the model's accuracy and F1-score. We observed that certain classes, such as Aspen Forest, are readily distinguishable due to their prevalence in the dataset. However, classes like Oak Forest and Beech Forest proved more challenging for the model to classify accurately. Additionally, we used cloud points derived from LiDAR images as a proxy for the ground truth height. Using this ground truth, it could be seen that the model tends to overestimate the height of certain types of trees. Our findings underscore the inherent difficulties in classification, particularly for forest types like Oak-Beech Forests, where moisture and dense vegetation complicate the disentanglement of ground and vegetation signals. This work opens the floor for new works using the P-band and the tomographic reconstruction for tree species classification.

\bibliographystyle{unsrtnat}
\bibliography{references}

\appendix

\section{Appendix}

\subsection{Dataset}\label{dataset_info}

\begin{table}[h!]
\caption{Forest Types that can be found in the area of study in the National park in Germany, its respective quantity, and the percentages of each one}
\label{tab:forest_types_with_percentages}
\centering
\begin{tabular}{|c|c|l|r|r|}
\hline
\textbf{Index} & \textbf{Code} & \textbf{Name} & \textbf{Count} & \textbf{Percentage} \\ \hline
1 & AA0 & Aspen forest & 64389 & 60.34\% \\ \hline
2 & AA1 & Pine forest & 5199 & 4.87\% \\ \hline
3 & AA2 & Beech forest with deciduous woods & 2884 & 2.70\% \\ \hline
4 & AB0 & Douglas fir forest & 11593 & 10.86\% \\ \hline
5 & AJ0 & Mixed spruce forest with native deciduous woods & 18120 & 16.98\% \\ \hline
6 & AJ1 & Oak-beech forest & 2590 & 2.43\% \\ \hline
7 & AK0 & Oak forest & 869 & 0.81\% \\ \hline
8 & AS0 & Beech forest & 1063 & 1.00\% \\ \hline
\end{tabular}
\end{table}

Table \ref{tab:forest_types_with_percentages} shows the pixel counts for each tree type in the study area within the national park in Germany. The dataset encompasses eight tree types. Aspen forest is the most dominant, representing 60\% of the data, followed by Mixed Spruce Forest. Oak Forest and Beech Forest each constitute less than 1\% of the total data.

\subsection{Model}

\begin{figure}
    \centering
        \caption{The figure is read from left to right. Initially, the tomographic image is converted into a tabular format with X and Y coordinates. The data is then allocated to its corresponding split, either swath or square. The data can be processed either by separating it by polarimetry channels or by combining them. Finally, this processed data is integrated with tree species classification information in a tabular format and submitted to AutoGluon for analysis.}
    \label{fig:model}
    \includegraphics[width=1.0\linewidth]{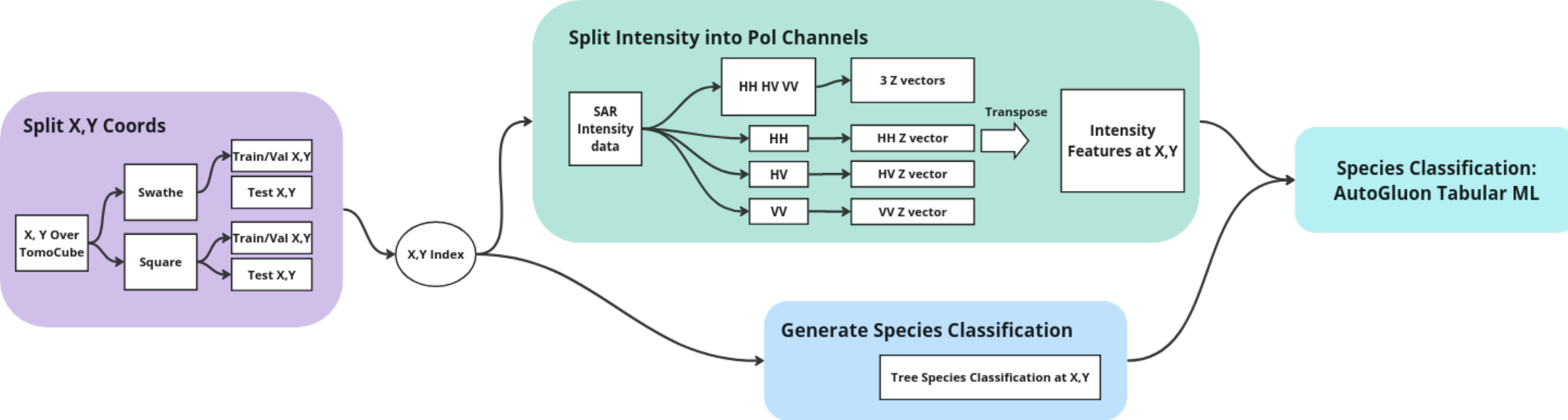}
\end{figure}

Fig. \ref{fig:model} presents the architecture proposed in this work.

\subsection{Classification Related to Height Estimation}
\begin{figure}
    \centering
        \caption{Each graph displays violin plots, with the kernel density estimation shown on both sides and the corresponding box plot in the center. The top left and bottom left corners depict the LiDAR heights for each actual tree class in the testing and training partitions, respectively. The top right and bottom right corners illustrate the LiDAR heights for each predicted tree class in the testing and training partitions, respectively.}
    \label{fig:violin_plots}
    \includegraphics[width=1\linewidth]{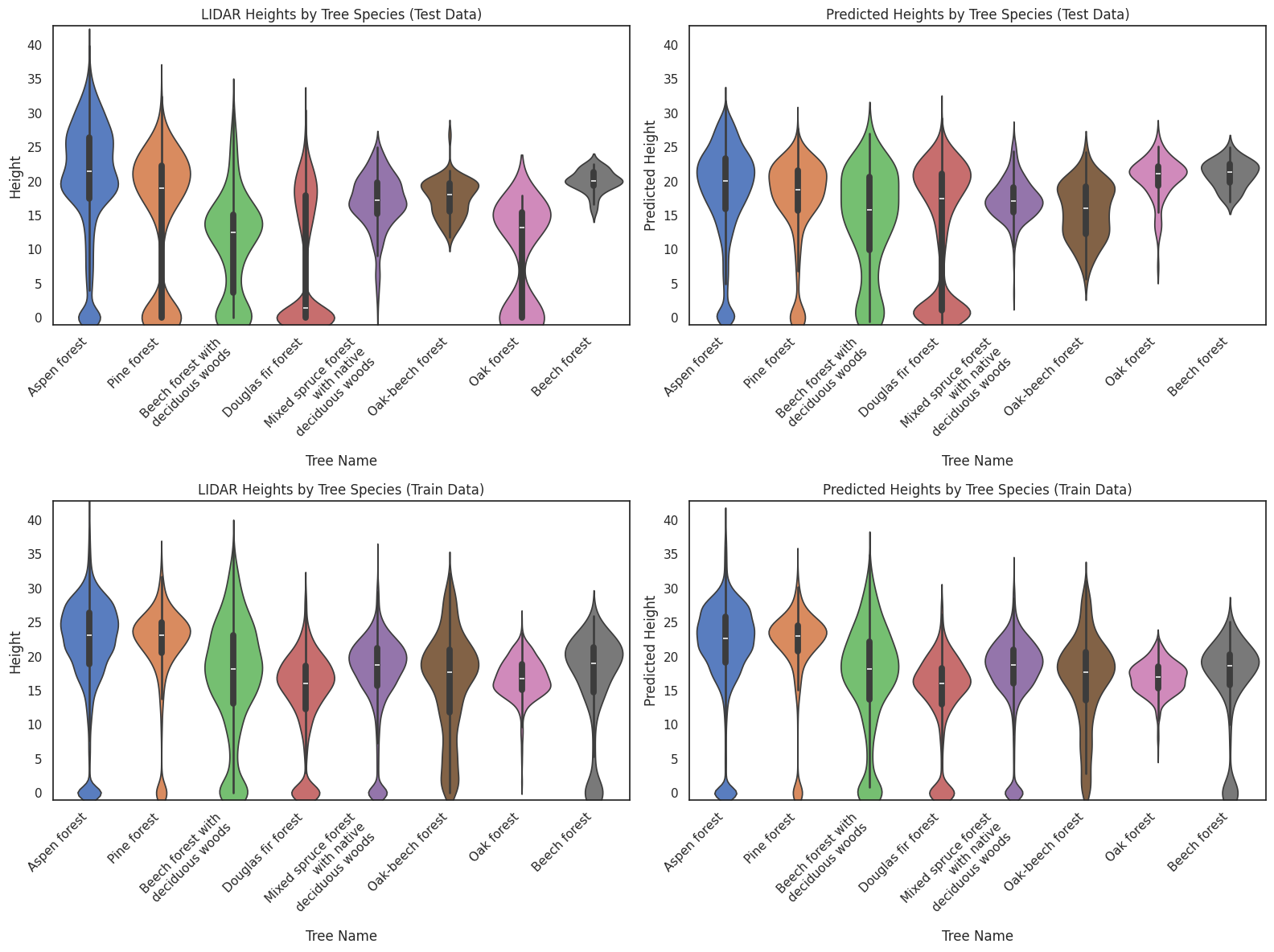}
\end{figure}

Fig. \ref{fig:violin_plots} displays the kernel density estimation and box plots for the proxy LiDAR data and the predicted classifications. We presented the figures for training and testing splits using the geo-split swath.

\begin{table}[h!]
\centering
\small
\begin{tabular}{|l|c|c|c|c|c|c|c|}
\hline
\textbf{Tree Name} & \textbf{Min (m)} & \textbf{Max (m)} & \textbf{Mean (m)} & \textbf{Std Dev (m)} & \textbf{Kurtosis} & \textbf{RMSE (m)} & \textbf{Split} \\ \hline
Aspen Forest & 0.36 & 42.52 & 23.35 & 5.68 & 0.81 & 5.28 & Test \\ \hline
Pine Forest & 8.77 & 27.67 & 22.61 & 2.59 & 1.87 & 3.75 & Test \\ \hline
\makecell[l]{Beech Forest with \\ Deciduous Woods} & 4.87 & 25.29 & 16.60 & 3.80 & 0.16 & 4.30 & Test \\ \hline
Douglas Fir Forest & 1.89 & 26.42 & 16.31 & 3.46 & 0.96 & 4.45 & Test \\ \hline
\makecell[l]{Mixed Spruce Forest with \\ Native Deciduous Woods} & 0.03 & 28.67 & 18.26 & 4.63 & 2.56 & 4.12 & Test \\ \hline
Oak-Beech Forest & 0.15 & 31.53 & 17.05 & 8.20 & -0.33 & 6.43 & Test \\ \hline
Oak Forest & 1.40 & 22.92 & 16.24 & 2.61 & 3.02 & 2.94 & Test \\ \hline
Beech Forest & 5.65 & 23.75 & 19.05 & 3.14 & 2.26 & 2.93 & Test \\ \hline \hline
Aspen Forest & 0.02 & 42.78 & 22.93 & 5.20 & 0.82 & 1.22 & Train \\ \hline
Pine Forest & 0.00 & 34.36 & 22.47 & 4.15 & 4.55 & 1.08 & Train \\ \hline
\makecell[l]{Beech Forest with \\ Deciduous Woods} & 0.44 & 36.45 & 19.59 & 6.66 & -0.06 & 1.37 & Train \\ \hline
Douglas Fir Forest & 0.43 & 30.47 & 17.13 & 3.97 & 0.50 & 1.13 & Train \\ \hline
\makecell[l]{Mixed Spruce Forest with \\ Native Deciduous Woods} & 0.29 & 34.75 & 18.77 & 4.12 & 1.98 & 1.23 & Train \\ \hline
Oak-Beech Forest & 0.19 & 32.22 & 16.46 & 6.88 & 2.03 & 1.53 & Train \\ \hline
Oak Forest & 1.90 & 25.21 & 17.17 & 3.06 & 2.75 & 1.05 & Train \\ \hline
Beech Forest & 1.54 & 25.92 & 19.01 & 4.19 & 1.72 & 1.26 & Train \\ \hline
\end{tabular}
\caption{LiDAR Height Statistics for the Test and Train Splits}
\label{tab:train_test_data_statistics}
\end{table}

\end{document}